\documentclass{article} 
\usepackage{nips15submit_e,times}
\usepackage{amsmath}
\usepackage{amsfonts}
\usepackage{hyperref}
\usepackage{url}
\usepackage{graphicx}
\usepackage{caption}
\usepackage{subcaption}

\DeclareMathOperator*{\argmin}{arg\,min}

\title{Discovering Hidden Factors of Variation in Deep Networks}

\author{
Brian Cheung\thanks{ Authors contributed equally.}\\
Redwood Center for Theoretical Neuroscience\\
University of California, Berkeley\\
Berkeley, CA 94720, USA\\
\texttt{bcheung@berkeley.edu}\\
\And
Jesse A. Livezey\footnotemark[1]\\
Redwood Center for Theoretical Neuroscience\\
University of California, Berkeley\\
Berkeley, CA 94720, USA\\
\texttt{jesse.livezey@berkeley.edu}\\
\AND
Arjun K. Bansal\\
Nervana Systems, Inc.\\
San Diego, CA 92121, USA\\
\texttt{arjun@nervanasys.com}\\
\And
Bruno A. Olshausen\\
Redwood Center for Theoretical Neuroscience\\
University of California, Berkeley\\
Berkeley, CA 94720, USA\\
\texttt{baolshausen@berkeley.edu}\\
}

\nipsfinalcopy 

\begin{document}

\maketitle

\begin{abstract}
Deep learning has enjoyed a great deal of success because of its ability to learn useful features for tasks such as classification. But there has been less exploration in learning the factors of variation apart from the classification signal. By augmenting autoencoders with simple regularization terms during training, we demonstrate that standard deep architectures can discover and explicitly represent factors of variation beyond those relevant for categorization. We introduce a cross-covariance penalty (XCov) as a method to disentangle factors like handwriting style for digits and subject identity in faces. We demonstrate this on the MNIST handwritten digit database, the Toronto Faces Database (TFD) and the Multi-PIE dataset by generating manipulated instances of the data. Furthermore, we demonstrate these deep networks can extrapolate `hidden' variation in the supervised signal.
\end{abstract}

\section{Introduction}
One of the goals of representation learning is to find an efficient representation of input data that simplifies tasks such as object classification \cite{krizhevsky2012imagenet} or image restoration \cite{eigen2013restoring}. Supervised algorithms approach this problem by learning features which transform the data into a space where different classes are linearly separable. However this often comes at the cost of discarding other variations such as style or pose that may be important for more general tasks. On the other hand, unsupervised learning algorithms such as autoencoders seek efficient representations of the data such that the input can be fully reconstructed, implying that the latent representation preserves all factors of variation in the data. However, without some explicit means for factoring apart the different sources of variation the factors relevant for a specific task such as categorization will be entangled with other factors across the latent variables.  Our goal in this work is to combine these two approaches to disentangle class-relevant signals from other factors of variation in the latent variables in a standard deep autoencoder.

Previous approaches to separating factors of variation in data, such as ‘content’ vs. ‘style’ \cite{tenenbaum2000separating} or ‘form’ vs. ‘motion’ \cite{grimes2005bilinear, olshausen2007bilinear, hinton2011transforming, memisevic2010learning, berkes2009structured, cadieu2012learning}, have relied upon a bilinear model architecture in which the units representing different factors are combined multiplicatively. Such an approach was recently utilized to separate facial expression vs. identity using higher-order restricted Boltzmann machines \cite{reed2014learning}. One downside of bilinear approaches in general is that they require learning an approximate weight tensor corresponding to all three-way multiplicative combinations of units. Despite the impressive results achieved with this approach, the question nevertheless remains as to whether there is a more straightforward way to separate factors of variation using standard nonlinearities in feedforward neural networks. Earlier work by \cite{salakhutdinov2007learning} demonstrated class-irrelevant aspects in MNIST (style) can be learned by including additional unsupervised units alongside supervised ones in an autoencoder. However, their model does not disentangle class-irrelevant factors from class-relevant ones. More recently, \cite{kingma2014semi} utilized a variational autoencoder in a semi-supervised learning paradigm which is capable of separating content and style in data. It is this work which is the inspiration for the simple training scheme presented here.

Autoencoder models have been shown to be useful for a variety of machine learning tasks \cite{rifai2011contractive, vincent2010stacked, le2013building}. The basic autoencoder architecture can be separated into an encoding stage and a decoding stage. During training, the two stages are jointly optimized to reconstruct the input data from the output of the decoder. In this work, we propose using both the encoding and decoding stages of the autoencoder to learn high-level representations of the factors of variation contained in the data. The high-level representation (or encoder output) is divided into two sets of variables. The first set (observed variables) is used in a discriminative task and during reconstruction. The second set (latent variables) is used only for reconstruction. To promote disentangling of representations in an autoencoder, we add two additional costs to the network. The first is a discriminative cost on the observed variables. The second is a novel cross-covariance penalty (XCov) between the observed and latent variables across a batch of data. This penalty prevents latent variables from encoding input variations due to class label. \cite{rifai2012disentangling} proposed a similar penalty over terms in the product between the Jacobians of observed and latent variables with respect to the input. In our penalty, the variables which represent class assignment are separated from those which are encoding other factors of variations in the data.

We analyze characteristics of this learned representation on three image datasets. In the absence of standard benchmark task for evaluating disentangling performance, our evaluation here is based on examining qualitatively what factors of variation are discovered for different datasets. In the case of MNIST, the learned factors correspond to style such as slant and size. In the case TFD the factors correspond to identity, and in the case of Multi-PIE identity specific attributes such as clothing, skin tone, and hair style. 

\section{Semi-supervised Autoencoder}

Given an input $\mathbf{x} \in \mathbb{R}^D$ and its corresponding class label $\mathbf{y} \in \mathbb{R}^L$ for a dataset $\mathcal{D}$, we consider the class label to be a high-level representation of its corresponding input. However, this representation is usually not invertible because it discards much of the variation contained in the input distribution. In order to properly reconstruct $\mathbf{x}$, autoencoders must learn a latent representation which preserves all input variations in the dataset.

\begin{figure}[h]
\begin{center}
\includegraphics[width=0.5\textwidth]{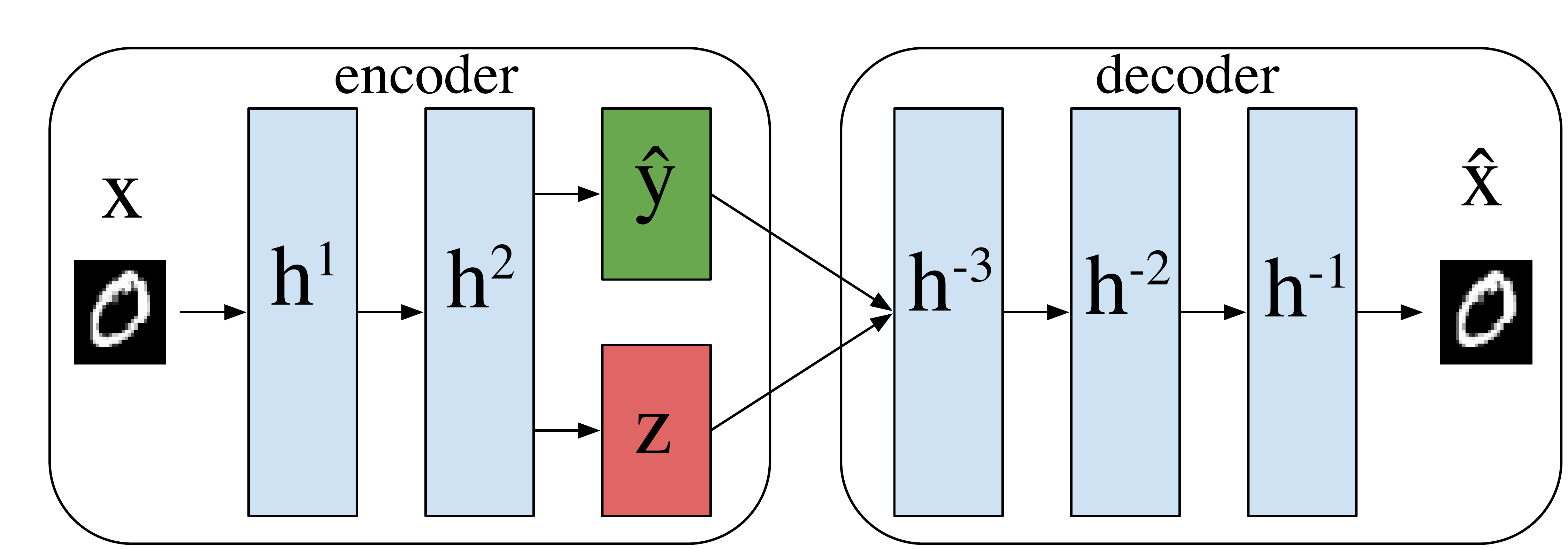}
\end{center}
\caption{The encoder and decoder are combined and jointly trained to reconstruct the inputs and predict the observed variables $\mathbf{\hat y}$.}
\label{fig:network}
\end{figure}

Using class labels, we incorporate supervised learning to a subset of these latent variables transforming them into observed variables, $\mathbf{\hat y}$ as shown in Figure \ref{fig:network}. In this framework, the remaining the latent variable $\mathbf{z}$ must account for the remaining variation of dataset $\mathcal{D}$. We hypothesize this latent variation is a high-level representation of the input complementary to the observed variation. For instance, the class label `5' provided by $\mathbf{y}$ would not be sufficient for the decoder to properly reconstruct the image of a particular `5'. In this scenario, $\mathbf{z}$ would encode properties of the digit such as style, slant, width, etc. to provide the decoder sufficient information to reconstruct the original input image. Mathematically, the encoder $F$ and decoder $G$ are defined respectively as:

\begin{align}
\{\mathbf{\hat y}, \mathbf{z}\} &= F(\mathbf{x}; \mathbf{\theta})\\
\mathbf{\hat{x}} &= G(\mathbf{y}, \mathbf{z}; \mathbf{\phi})
\end{align}

where $\mathbf{\theta}$ and $\mathbf{\phi}$ are the parameters of the encoder and decoder respectively. 

\subsection{Learning}
The objective function to train the network is defined as the sume of three seperate cost terms.

\begin{equation}
\label{eq:costs}
\mathbf{\hat{\theta}},\mathbf{\hat{\phi}} = 
\argmin_{\mathbf{\theta},\mathbf{\phi}} \sum_{\{\mathbf{x},\mathbf{y}\} \in \mathcal{D}} 
||\mathbf{x} - \mathbf{\hat x}||^2 
+ \beta \sum_{i} y_i log(\hat y_i)
+ \gamma C.
\end{equation}

The first term is a typical reconstruction cost (squared error) for an autoencoder. The second term is a standard supervised cost (cross-entropy). While there are many potential choices for the reconstruction cost depending on the distribution of data vector $x$, for our experiments we use squared-error for all datasets. For the observed variables, the form of the cost function depends on the type of variables (categorical, binary, continuous). For our experiments, we had categorical observed variables so we parametrized them as one-hot vectors and compute $\mathbf{\hat y} = softmax(\mathbf{W_{\hat y} h^2} + \mathbf{b_{\hat y}})$.

The third term $C$ is the unsupervised cross-covariance (XCov) cost which disentangles the observed and latent variables of the encoder.

\begin{equation}
\label{eq:xcov_cost}
C(\mathbf{\hat y}^{1...N}, \mathbf{z}^{1...N}) = \frac{1}{2} \sum_{ij} [\frac{1}{N}\sum_n(\hat y^n_i- \bar{\hat{y}}_i)(z^n_j-\bar{\hat{z}}_j)]^2.
\end{equation}

The XCov penalty to disentangle $\mathbf{\hat y}$ and $\mathbf{z}$ is simply a sum-squared cross-covariance penalty between the activations across samples in a batch of size $N$ where $\bar{\hat{y}}_i$ and $\bar{\hat{z}}_j$ denote means over examples. $n$ is an index over examples and $i,j$ index feature dimensions. Unlike the reconstruction and supervised terms in the objective, XCov is a cost computed over a batch of datapoints. It is possible to approximate this quantity with a moving average during training but we have found that this cost has been robust to small batch sizes and have not found any issues when training with mini-batches as small as $N = 50$. Its derivative is provided in the supplementary material.

This objective function naturally fits a semi-supervised learning framework. For unlabeled data, the multiplier $\beta$ for the supervised cost is simply set to zero. In general, the choice of $\beta$ and $\gamma$ will depend on the intended task. Larger $\beta$ will lead to better to classification performance while larger $\gamma$ to better separation between latent and observed factors.

\section{Experimental Results}

We evaluate autoencoders trained to minimize \ref{eq:costs} on three datasets of increasing complexity. The network is trained using ADADELTA \cite{zeiler2012adadelta} with gradients from standard backpropagation. Models were implemented in a modified version of Pylearn2 \cite{goodfellow2013pylearn2} using deconvolution and likelihood estimation code from \cite{goodfellow2014generative}.

\subsection{Datasets}

\subsubsection*{MNIST Handwritten Digits Database}
The MNIST handwritten digits database \cite{lecun1998mnist} consists of 60,000 training and 10,000 test images of handwritten digits 0-9 of size 28x28. Following previous work \cite{goodfellow2013maxout}, we split the training set into 50,000 samples for training and 10,000 samples as a validation set for model selection. 

\subsubsection*{Toronto Faces Database}
\begin{figure}[h]
\centering
\begin{subfigure}{.5\textwidth}
\centering
\includegraphics[width=0.65\linewidth]{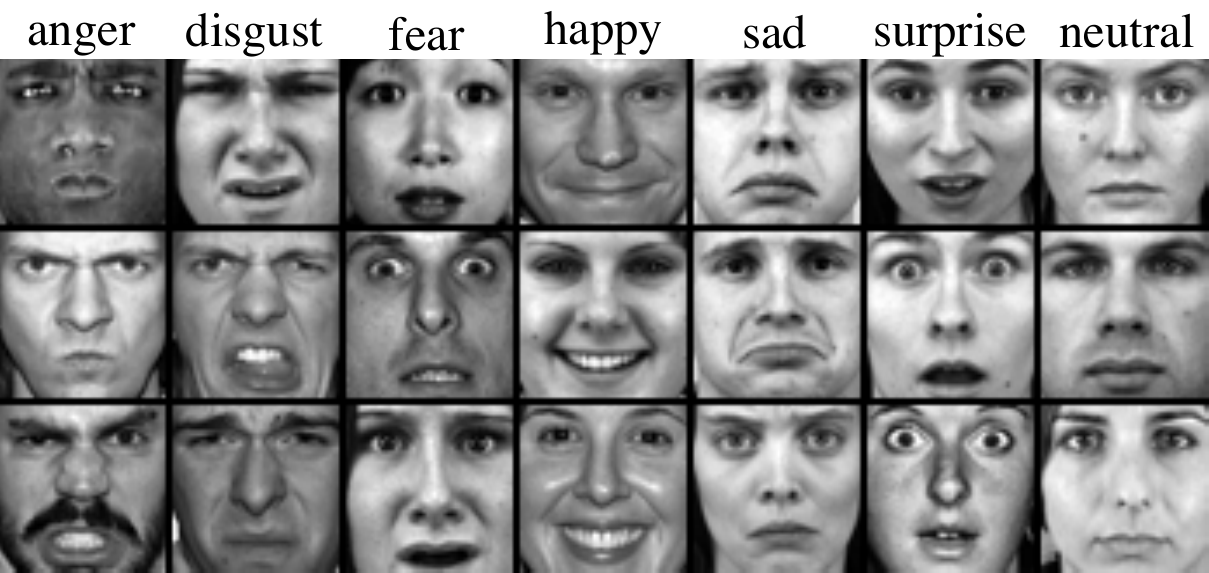}
\end{subfigure}%
\begin{subfigure}{.5\textwidth}
\centering
\includegraphics[width=1.0\linewidth]{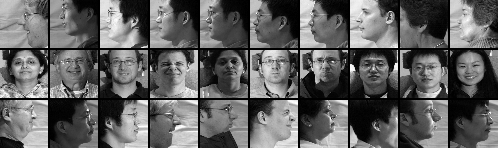}
\end{subfigure}
\caption{Left: Example TFD images from the test set showing 7 expressions with random identity. Right: Example Multi-PIE images from the test set showing 3 of the 19 camera poses with variable lighting and identity.}
\label{fig:tfd_mpie_examples}
\end{figure}

The Toronto Faces Database \cite{tfd} consists of 102,236 grayscale face images of size 48x48. Of these, 4,178 are labeled with 1 of 7 different expressions (anger, disgust, fear, happy, sad, surprise, and neutral). Examples are shown in Figure \ref{fig:tfd_mpie_examples}. The dataset also contains 3,784 identity labels which were not used in this paper. The dataset has 5 folds of training, validation and test examples. The three partitions are disjoint and contain no overlap in identities.

\subsubsection*{Multi-PIE Dataset}
The Multi-PIE datasets \cite{gross2010multi} consists of 754,200 high-resolution color images of 337 subjects. Each subject was recorded under 15 camera poses: 13 spaced at 15 degree intervals at head height, and 2 positioned above the subject. For each of these cameras, subjects were imaged under 19 illumination conditions and a variety of facial expressions. We discarded images from the two overhead cameras due to inconsistencies found in their image. Camera pose and illumination data was retained as supervised labels.

Only a small subset of the images possess facial keypoint information for each camera pose. To perform a weak registration to appoximately localize the face region, we compute the maximum bounding box created by all available facial keypoint coordinates for a given camera pose. This bounding box is applied to all images for that camera pose. We then resized the cropped images to 48x48 pixels and convert to grayscale. We divide the dataset into 528,060 training, 65,000 validation and 60,580 test examples. Splits were determined by subject id. Therefore, the test set contains no overlap in identities with the training or validation sets. Example images from our test set are shown in Figure \ref{fig:tfd_mpie_examples}.

The Multi-PIE dataset contains significantly more complex factors of variation than MNIST or TFD. Unlike TFD, images in Multi-PIE includes much more of the subject's body. The weak registration also causes significant variation in the subject's head position and scale.

\begin{table}[h]
\caption{Network Architectures (Softmax (SM), Rectified Linear (ReLU))}
\label{table:architectures}
\begin{center}
\begin{tabular}{lll}
\textbf{MNIST}  & \textbf{TFD}     & \textbf{ConvDeconvMultiPIE} \\
\hline \\
500 ReLU        & 2000 ReLU        & 20x20x32 ConvReLU           \\
500 ReLU        & 2000 ReLU        & 2000 ReLU                   \\
10 SM, 2 Linear & 7 SM, 793 Linear & 2000 ReLU                   \\
500 ReLU        & 2000 ReLU        & 13 SM, 19 SM, 793 Linear    \\
500 ReLU        & 2000 ReLU        & 2000 ReLU                   \\
784 Linear      & 2304 Linear      & 2000 ReLU                   \\
                &                  & 2000 ReLU                   \\
                &                  & 2000 ReLU                   \\
                &                  & 48x48x1 Deconv             
\end{tabular}
\end{center}
\end{table}

\subsection{Model Performace}

\subsubsection{Classification}
As a sanity check, we first show that the our additional regularization term in the cost negligibly impacts performance for different architectures including convolution, maxout and dropout. Tables \ref{table:class_mnist} and \ref{table:class_tfd} show classification results for MNIST and TFD are comparable to previously published results. Details on network architecture for these models can be found in the supplementary material.

\begin{table}[t]
\caption{MNIST Classification Performance}
\label{table:class_mnist}
\begin{center}
\begin{tabular}{lll}
\multicolumn{1}{c}{\bf Model} &\multicolumn{1}{c}{\bf Accuracy}&\multicolumn{1}{c}{\bf Model Selection Criterion}
\\ \hline \\
MNIST   & 98.35 & Reconstuction: $\beta=10,\gamma=10$\\
ConvMNIST   & 98.71 & Reconstuction: $\beta=10,\gamma=10$\\
MaxoutMNIST + dropout & 99.01 & Accuracy: $\beta=100,\gamma=10$ \\
Maxout + dropout \cite{goodfellow2013maxout} & 99.06 & Accuracy\\
\end{tabular}
\end{center}
\end{table}

\begin{table}[t]
\caption{TFD Classification Performance}
\label{table:class_tfd}
\begin{center}
\begin{tabular}{lll}
\multicolumn{1}{c}{\bf Model} &\multicolumn{1}{c}{\bf Accuracy}&\multicolumn{1}{c}{\bf Model Selection Criterion}
\\ \hline \\
TFD & 69.4 & Reconstuction: $\beta=10,\gamma=1e3$ (Fold 0) \\
ConvTFD & 84.0 & Accuracy: $\beta=10,\gamma=1e3$ (Fold 0) \\
disBM \cite{reed2014learning} & 85.4 & Accuracy\\
CCNET+CDA+SVM \cite{rifai2012disentangling} & 85.0 & Accuracy (Fold 0)\\
\end{tabular}
\end{center}
\end{table}

\subsubsection{Learned Factors of Variation}

\begin{figure}[h]
\begin{center}
\includegraphics[width=1.\textwidth]{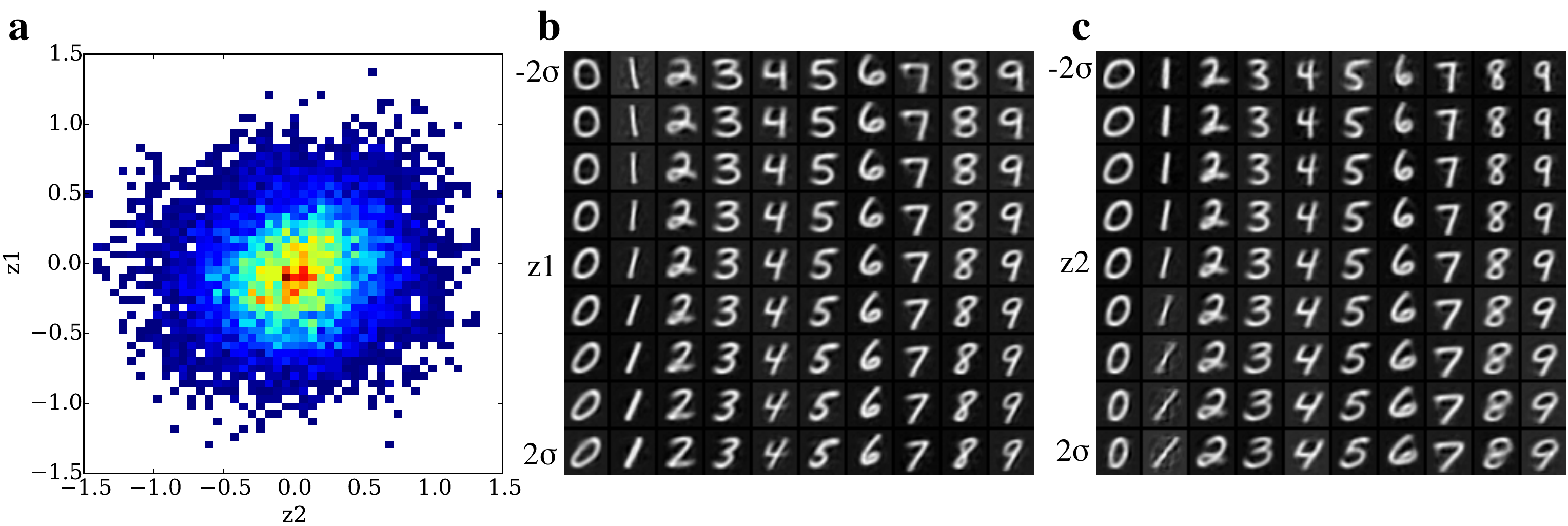}
\end{center}
\caption{\textbf{a}: Histogram of test set $z$ variables. \textbf{b}: Generated MNIST digits formed by setting $z_2$ to zero and varying $z_1$. \textbf{c}: Generated MNIST digits formed by setting $z_1$ to zero and varying $z_2$. $\sigma$ was calculated from the variation on the test set.}
\label{fig:mnist_z}
\end{figure}

We begin our analysis using the MNIST dataset. We intentionally choose $\mathbf{z} \in \mathbb{R}^2$ for the architecture described in Table \ref{table:architectures} for ease in visualization of the latent variables. As shown in Figure \ref{fig:mnist_z}a, $\mathbf{z}$ takes on a suprisingly simple isotropic Normal distribution with mean 0 and standard deviation .35.

\noindent\textit{Visualizing Latent Variables}

To visualize the transformations that the latent variables are learning, the decoder can be used to create images for different values of $\mathbf{z}$. We vary a single element $z_i$ linearly over a set interval with $z_{\setminus{i}}$ fixed to 0 and $\mathbf{y}$ fixed to one-hot vectors corresponding to each class label as shown in Figure \ref{fig:mnist_z}b and c. Moving across each column for a given row, the digit style is maintained as the class labels varies. This suggests the network has learned a class invariant latent representation. At the center of $\mathbf{z}$-space, (0,0), we find the canonical MNIST digit style. Moving away from the center, the digits become more stylized but also less probable. We find this result is reliably reproduced without the XCov regularization when the dimensionality of $\mathbf{z}$ is relatively small suggesting that the network naturally prefers such a latent representation for factors of variation absent in the supervised signal. With this knowledge, we describe a method to generate samples from such an autoencoder with competative generative performance in the supplementary material.

\noindent\textit{Moving From Latent Space to Image Space}

\begin{figure}[h]
\begin{center}
\includegraphics[width=0.85\textwidth]{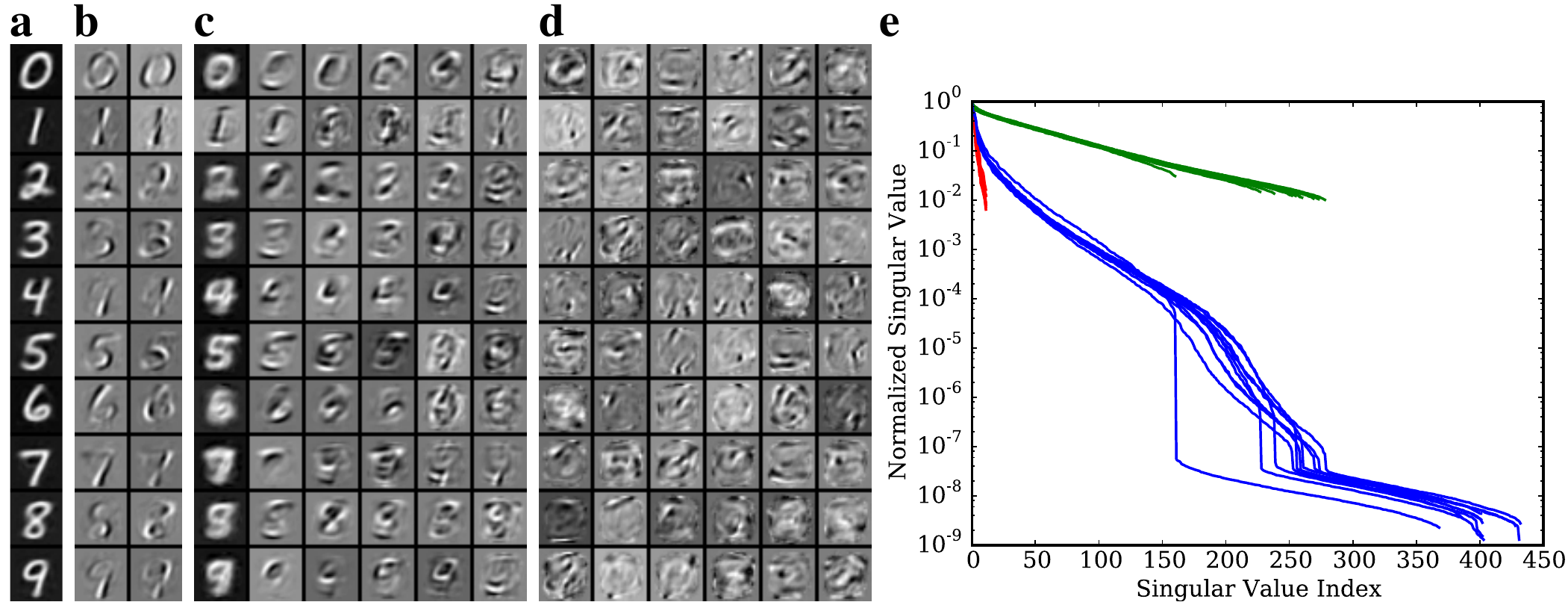}
\end{center}
\caption{\textbf{a}: Jacobians were taken at activation values that lead to these images. $\mathbf{z}$ was set to zero for each digit class. \textbf{b}: Gradients of the decoder output with respect to $\mathbf{z}$. \textbf{c}: Singular vectors from the Jacobian from the activations of the first layer after \{$\mathbf{y}$, $\mathbf{z}$\}. \textbf{d}: Column vectors from the Jacobian from the the activations of the second layer after \{$\mathbf{y}$, $\mathbf{z}$\}. Note that units in the columns for \textbf{d} are not neccesarily the same unit. \textbf{e}: Plots of the normalized singular values for \textit{red}: the Jacobians with respect to \{$\mathbf{y}$, $\mathbf{z}$\}, \textit{blue}: the Jacobians with respect to the activations of the first layer after \{$\mathbf{y}$, $\mathbf{z}$\} ($\mathbf{h^{-3}}$ in Figure \ref{fig:network}), and \textit{green}: the Jacobians with respect to the activations of the second layer after \{$\mathbf{y}$, $\mathbf{z}$\} ($\mathbf{h^{-2}}$ in Figure \ref{fig:network}).}
\label{fig:mnist_grads}
\end{figure}

Following the layer of observed and latent variables \{$\mathbf{y}$, $\mathbf{z}$\}, there are two additional layers of activations $\mathbf{h^{-3}}, \mathbf{h^{-2}}$ before the output of the model into image space. To visualize the function of these layers, we compute the Jacobian of the output image, $\mathbf{\hat x}$, with respect to the activation of hidden units, $\mathbf{h^k}$, in a particular layer. This analysis provides insight into the transformation each unit is applying to the input $\mathbf{x}$ to generate $\mathbf{\hat x}$. More specifically, it is a measure of how a small perturbation of a particular unit in the network affects the output $\mathbf{\hat x}$:
\begin{equation}
\Delta \hat x^k_{i} = \frac{\partial\hat x_i}{\partial h^k_j}\Delta h_j.
\end{equation}
Here, $i$ is the index of a pixel in the output of the network, $j$ is the index of a hidden unit, and $k$ is the layer number.
We remove hidden units with zero activation from the Jacobian since their derivatives are not meaningful. A summary of the results are plotted in Figure \ref{fig:mnist_grads}.

The Jacobian with respect to the $\mathbf{z}$ units shown in Figure \ref{fig:mnist_grads}b locally mirror the transformations seen in Figure \ref{fig:mnist_z}b and c further confirming the hypothesis that this latent space smoothly controls digit style. The slanted style generated as $z_2$ approaches $2\sigma$ in Figure \ref{fig:mnist_z}c is created by applying a gabor-like filter to vertically oriented parts of the digit as shown in the second column of Figure \ref{fig:mnist_grads}b.

Rather than viewing each unit in the next layer individually, we analyze the singular value spectrum of the Jacobian. For $\mathbf{h^{-3}}$, the spectrum is peaked and thus there are a small number of directions with large effect on the image output, so we plot singular vectors with largest singular value. For all digits besides `1', the first component seems to create a template digit and the other componets make small style adjustments. For $\mathbf{h^{-2}}$, the spectrum is more degenerate, so we choose a random set of columns from the Jacobian to plot which will better represent the layer's function. We notice that for each layer moving from the encoder to the output, their contributions become more spatially localized and less semantically meaningful.

\subsection{Generating Expression Transformations}
\begin{figure}[h]
\begin{center}
\includegraphics[width=0.75\textwidth]{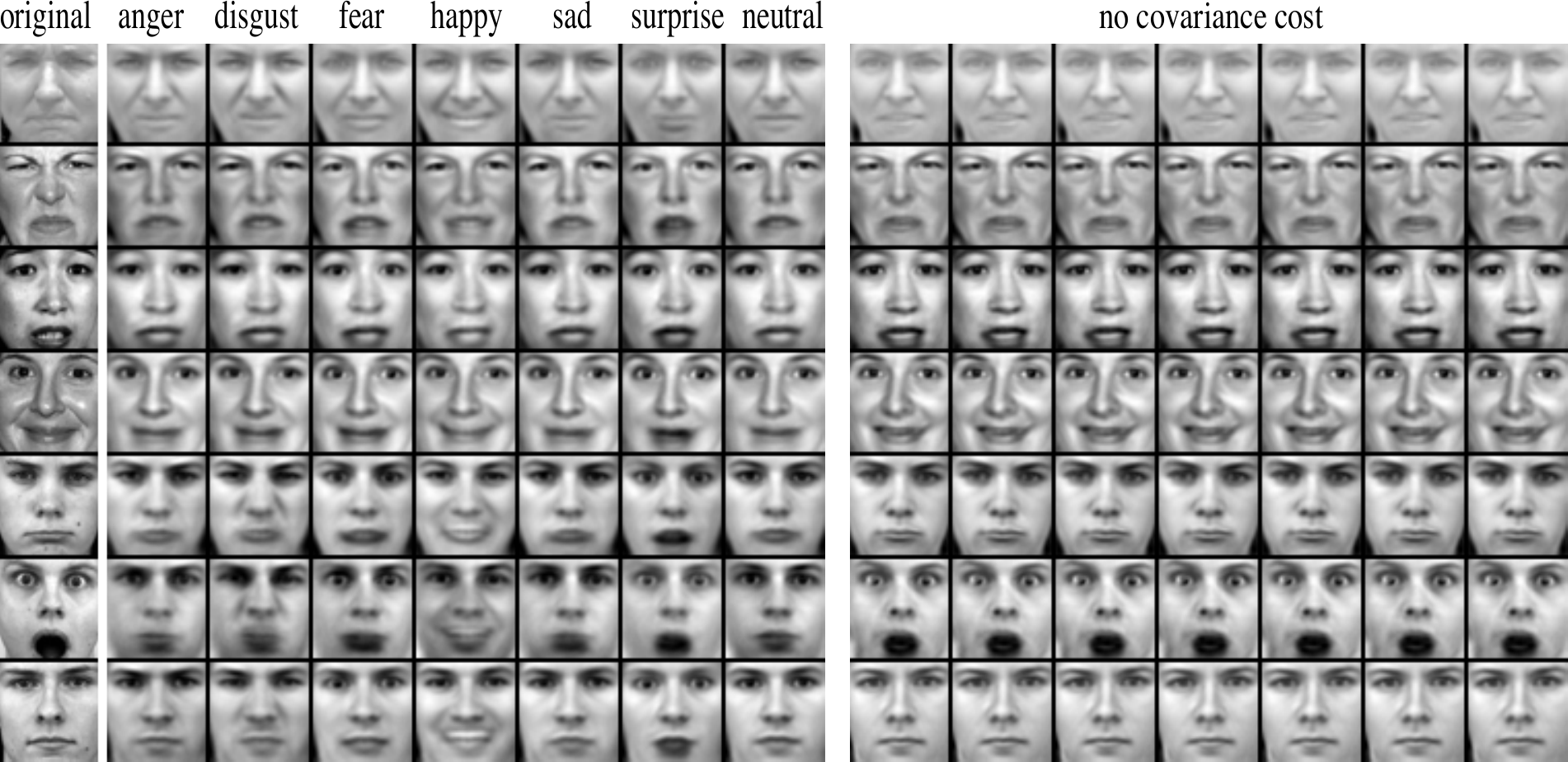}
\end{center}
\caption{Left column: Samples from the test set displaying each of the 7 expressions. The expression-labeled columns are generated by keeping the latent variables $\mathbf{z}$ constant and changing $\mathbf{y}$ (expression). The rightmost set of faces are from a model with no covarriance cost and showcase the importance of the cost in disentangling expression from the latent $\mathbf{z}$ variables.}
\label{fig:tfd}
\end{figure}

We demonstrate similar manipulations on the TFD dataset which contains substantially more complex images than MNIST and has far fewer labeled examples. After training, we find the latent representation $\mathbf{z}$ encodes the subject's identity, a major factor of variation in the dataset which is not represented by the expression labels. The autoencoder is able to change the expression while preserving identity of faces never before seen by the model. We first initialize $\{\mathbf{\hat y}, \mathbf{z}\}$ with an example from the test set. We then replace $\mathbf{\hat y}$ with a new expression label $\mathbf{\hat y'}$ feeding $\{\mathbf{\hat y'}, \mathbf{z}\}$ to the decoder. Figure \ref{fig:tfd} shows the results of this process. Expressions can be changed while leaving other facial features largely intact. Similar to the MNIST dataset, we find the XCov penalty is not necessary when the dimensionality of $\mathbf{z}$ is low ($<$10). But convergence during training becomes far more difficult with such a bottleneck. We achieve much better reconstruction error with the XCov penalty and a high-dimensional $\mathbf{z}$. The XCov penalty simply prevents expression label variation from `leaking' into the latent representation. Figure \ref{fig:tfd} shows the decoder is unaffected by changes in $\mathbf{y}$ without the XCov penalty because the expression variation is distributed across the hundreds of dimensions of $\mathbf{z}$.

\subsection{Extrapolating Observed Variables}
\begin{figure}[h]
\begin{center}
\includegraphics[width=0.4\textwidth]{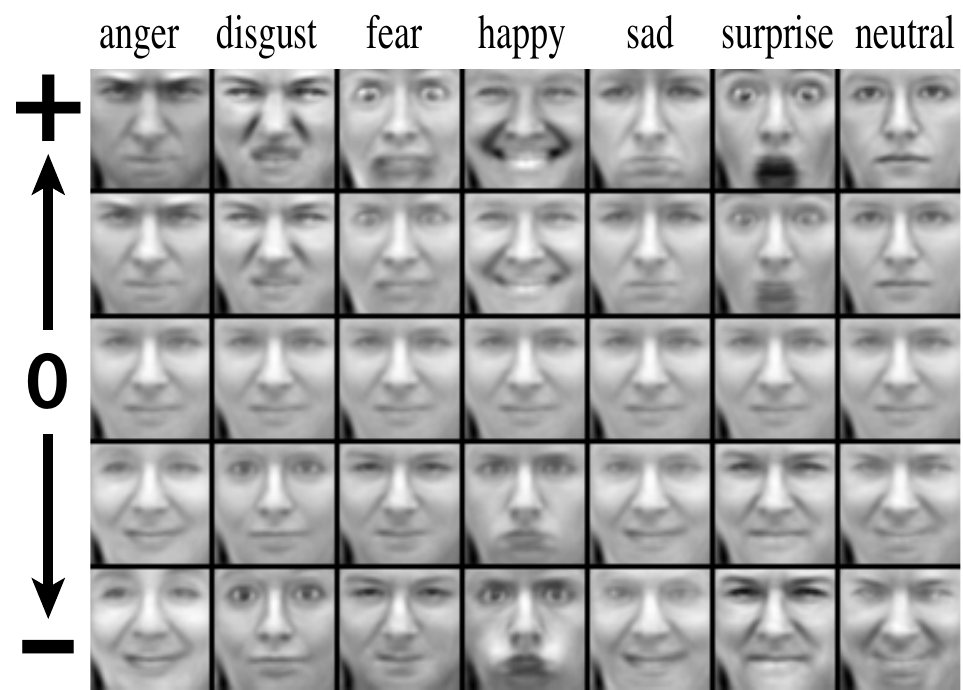}
\end{center}
\caption{For each column, $\mathbf{y}$ is set to a one-hot vector and scaled from 5 to -5 from top to bottom, well outside of the natural range of [0,1]. `Opposite' expressions and more extreme expressions can be made.}
\label{fig:tfd_spread}
\end{figure}

Previously, we showed the autoencoder learns a smooth continuous latent representation. We find a similar result for the observed expression variables despite only being provided their discrete class labels. In Figure \ref{fig:tfd_spread}, we go a step further. We try values for $\mathbf{\hat y}$ well beyond those that the encoder could ever output with a softmax activation (0 to 1). We vary the expression variable given to the decoder from -5 to 5. This results in greatly exagerated expressions when set to extreme positive values as seen in Figure \ref{fig:tfd_spread}. Remarkably, setting the variables to extreme negative values results in `negative` facial expressions being displayed. These negative facial expressions are abstract opposites of their positive counterparts. When the eyes are open in one extreme, they are closed in the opposite extreme. This is consistent regardless of the expression label and holds true for other abstract facial features such as open/closed mouth and smiling/frowning face. The decoder has learned a meaningful extrapolation of facial expression structure not explicitly present in the labeled data, creating a smooth semantically sensible space for values of the observed variables completely absent from the class labels.

\subsection{Manipulating Multiple Factors of Variation}

For Multi-PIE, we use two sets of observed factors (camera pose and illumination). As shown in Table \ref{table:architectures}, we have two softmax layers at the end of the encoder. The first encodes the camera pose of the input image and the second the illumination condition. Due to the increased complexity of these images, we made this network substantially deeper (9 layers).

\begin{figure}[h]
\begin{center}
\includegraphics[width=0.75\textwidth]{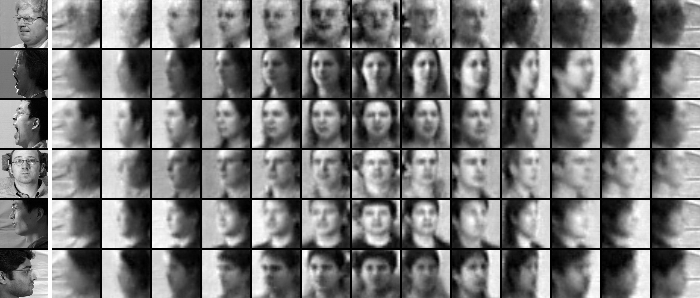}
\end{center}
\caption{Left column: Samples from test set with initial camera pose. The faces on the right were generated by changing the corresponding camera pose.}
\label{fig:mtpie_pose}
\end{figure}

\begin{figure}[h]
\begin{center}
\includegraphics[width=1.\textwidth]{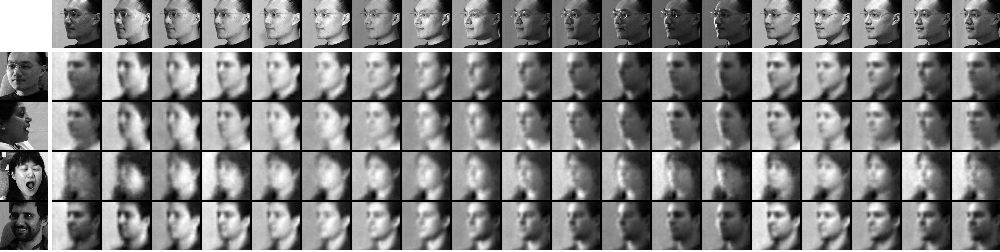}
\end{center}
\caption{Left column: Samples from test set. Illumination transformations are shown to the right. Ground truth lighting for the first face in each block is in the first row.}
\label{fig:mtpie_light}
\end{figure}

In Figure \ref{fig:mtpie_pose}, we show the images generated by the decoder while iterating through each camera pose. The network was tied to the illumination and latent variables of images from the test set. Although blurry, the generated images preserve the subject's illumination and identity (i.e. shirt color, hair style, skin tone) as the camera pose changes. In Figure \ref{fig:mtpie_light}, we instead fix the camera position and iterate through different illumination conditions. We also find it possible to interpolate between camera and lighting positions by simply linearly interpolating the $\mathbf{\hat y}$ between two neighboring camera positions supporting the inherent continuity in the class labels.

\section{Conclusion}


With the addition of a supervised cost and an unsupervised cross-covariance penalty, an autoencoder can learn to disentangle various transformations using standard feedforward neural network components. The decoder implicitly learns to generate novel manipulations of images on multiple sets of transformation variables. We show deep feedforward networks are capable of learning higher-order factors of variation beyond the observed labels without the need to explicitly define these higher-order interactions. Finally, we demonstrate the natural ability of these deep networks to learn a continuum of higher-order factors of variation in both the latent and observed variables. Surprisingly, these networks can extrapolate intrinsic continuous variation hidden in discrete class labels. These results gives insight in the potential of deep learning for the discovery of hidden factors of variation simply by accounting for known variation. This has many potential applications in exploratory data analysis and signal denoising.

\subsubsection*{Acknowledgments}
We would like to acknowledge everyone at the Redwood Center for their helpful discussion and comments. We thank Nervana Systems for supporting Brian Cheung during the summer when this project originated and for their continued collaboration. We gratefully acknowledge the support of NVIDIA Corporation with the donation of the Tesla K40 GPUs used for this research. Bruno Olshausen was supported by NSF grant IIS-1111765.


\bibliographystyle{IEEEbib}
\scriptsize{\bibliography{nips2015}}

\end{document}


\maketitle

\section{XCov Derivative}
\begin{align}
\label{eq:dcost}
C(\hat y, z) &= \frac{1}{2} \sum_{ij} [\frac{1}{N}\sum_n(\hat y^n_i- \bar{\hat{y}}_i)(z^n_j-\bar{\hat{z}}_j)]^2\\
\frac{\partial C(\hat y, z)}{\partial \hat y^m_a} &= \sum_j [\frac{1}{N}\sum_n(\hat y^n_a-\bar{\hat{y}}_a)(\hat z^n_j-\bar{\hat{z}}_j)]
[\frac{1}{N}(\hat z^m_j-\bar{\hat{z}}_j)]\\
\frac{\partial C(\hat y, z)}{\partial \hat z^m_b} &= \sum_i [\frac{1}{N}\sum_n(\hat y^n_i-\bar{\hat{y}}_i)(\hat z^n_b-\bar{\hat{z}}_b)]
[\frac{1}{N}(\hat y^m_i-\bar{\hat{y}}_i)]
\end{align}

\section{Network Architectures for Classification Performance}

\begin{table}[h]
\caption{Network Architectures (Softmax (SM), Rectified Linear (ReLU))}
\label{table:add_architectures}
\begin{center}
\begin{tabular}{lll}
\textbf{ConvMNIST} & \textbf{MaxoutMNIST} & \textbf{ConvTFD}  \\
\hline \\
12x12x32 ConvReLU  & 240-5 Maxout         & 22x22x16 ConvReLU \\
500 ReLU           & 240-5 Maxout         & 2000 ReLU         \\
10 SM, 2 Linear    & 10 SM, 2 Linear      & 7 SM, 793 Linear  \\
500 ReLU           & 240-5 Maxout         & 2000 ReLU         \\
500 ReLU           & 240-5 Maxout         & 2000 ReLU         \\
784 Linear         &   784 Linear                   & 2304 Linear       \\
                   &                      &                   \\
                   &                      &                   \\
                   &                      &                  
\end{tabular}
\end{center}
\end{table}

\section{Generative Performace}
Table \ref{table:generative} shows Parzen-window based likelihood estimates from our models introduced in \cite{breuleux2011quickly}. We use the code and results from \cite{goodfellow2014generative}. For our model, we first generate $y$s and $z$s from encoded training data. Then we fit a \{Categorical, Normal\} model to these latent variables and generate samples from this distribution and pass them into the decoder. We fit a Parzen-window model to these samples and evaluate the likelihood of the test set under this model. Cross validation of $\sigma$ was done in the same way as \cite{goodfellow2014generative}. We do not report state-of-the-art results, but our results are competitive with many of the previous methods.

We also estimate an empircal upper bound by fitting the Parzen-window model on actual training examples and evaluating the likelihood of the test set.

It is important to note that the choice of a Normal distrubution for $z$s is empirically and not theoretically motivated. We also artificially restricted the total latent space to 12 dimensions for the MNIST model to allow visualization.

\begin{table}[t]
\caption{Generative Performance}
\label{table:generative}
\begin{center}
\begin{tabular}{lll}
\multicolumn{1}{c}{\bf Model} &\multicolumn{1}{c}{\bf MNIST}&\multicolumn{1}{c}{\bf TFD}
\\ \hline \\
Table \ref{table:architectures} Models & 138 $\pm$ 1.6 &1913 $\pm$ 22\\
DBN \cite{bengio2013better} & 138 $\pm$ 2 & 1909 $\pm$ 66\\
Stacked CAE \cite{bengio2013better} & 121 $\pm$ 1.6 & \textbf{2110 $\pm$ 50}\\
Deep GSN \cite{bengio2014deep} & 214 $\pm$ 1.1 & 1890 $\pm$ 29\\
Adversarial nets \cite{goodfellow2014generative} & \textbf{225 $\pm$ 2}& \textbf{2057 $\pm$ 26}\\
\hline
Empirical Bound & 241 $\pm$ 2.1& 2129 $\pm$ 21\\
\end{tabular}
\end{center}
\end{table}

\bibliographystyle{IEEEbib}
\scriptsize{\bibliography{nips2015}}